\documentclass[11pt]{article}
\usepackage{authblk}

\newlength{\defbaselineskip}
\usepackage{graphicx}
\usepackage{amsmath}
\usepackage{amssymb}
\usepackage[caption=false,font=footnotesize]{subfig}
\usepackage{stmaryrd}

\usepackage{algorithmic}
\usepackage{algorithm}

\begin{document}

\title{
Kolmogorov-Arnold Networks in Low-Data Regimes: A Comparative Study with Multilayer Perceptrons
}

\author{Farhad Pourkamali-Anaraki
  \\
  \textit{Department of Mathematical and Statistical Sciences, University of Colorado Denver, CO, USA}
}

\date{\vspace{-7ex}}

\maketitle

\begin{abstract}
Multilayer Perceptrons (MLPs) have long been a cornerstone in deep learning, known for their capacity to model complex relationships. Recently, Kolmogorov-Arnold Networks (KANs) have emerged as a compelling alternative, utilizing highly flexible learnable activation functions directly on network edges, a departure from the neuron-centric approach of MLPs. However, KANs significantly increase the number of learnable parameters, raising concerns about their effectiveness in data-scarce environments. This paper presents a comprehensive comparative study of MLPs and KANs from both algorithmic and experimental perspectives, with a focus on low-data regimes. We introduce an effective technique for designing MLPs with unique, parameterized activation functions for each neuron, enabling a more balanced comparison with KANs. Using empirical evaluations on simulated data and two real-world data sets from medicine and engineering, we explore the trade-offs between model complexity and accuracy, with particular attention to the role of network depth. Our findings show that MLPs with individualized activation functions achieve significantly higher predictive accuracy with only a modest increase in parameters, especially when the sample size is limited to around one hundred. For example, in a three-class classification problem within additive manufacturing, MLPs achieve a median accuracy of 0.91, significantly outperforming KANs, which only reach a median accuracy of 0.53 with default hyperparameters. These results offer valuable insights into the impact of activation function selection in neural networks.
\end{abstract}

\section{Introduction}\label{sec:intro}
Multilayer Perceptrons (MLPs) are essential to modern deep learning due to their ability to model intricate nonlinear relationships \cite{raghu2017expressive}. MLPs consist of interconnected layers of neurons, forming a network where information flows from input to output. The connections between neurons in adjacent layers are represented by edges, each associated with a weight that determines the strength of the signal passed between them. Within each hidden layer, every neuron performs two key tasks. The first one is to calculate a weighted sum of its inputs, aggregating the signals received from the neurons in the previous layer. The second task involves applying a nonlinear activation function to this weighted sum. Without the nonlinear transformations provided by activation functions, neural networks would be limited to simple linear relationships, severely hindering their ability to extract complex patterns \cite{hayou2019impact}.

Traditionally, MLPs relied on fixed activation functions such as the Rectified Linear Unit (ReLU) \cite{vershynin2020memory} or the hyperbolic tangent (tanh). Training in this paradigm focuses mainly on adjusting edge weights to reduce error, keeping the activation functions constant. Recently, there has been a growing interest in parameterized activation functions, which offer greater control during the learning process \cite{agostinelli2014learning,lee2022stochastic}. These functions contain adjustable or learnable parameters, allowing the network to fine-tune its nonlinearities as it trains. This enhanced flexibility can improve performance by tailoring the model's internal transformations to the unique characteristics of the data.

In the pursuit of adaptive activation functions, significant research has focused on parameterizing functions that build upon ReLU, allowing for more nuanced handling of negative inputs. For instance, Leaky ReLU introduces a slight linear slope for negative values \cite{maniatopoulos2021learnable}, while the Exponential Linear Unit (ELU) employs an exponential function in this region \cite{trottier2017parametric}. Additionally, the Sigmoid Linear Unit (SiLU), also known as the swish function \cite{ramachandran2017searching,tanaka2020weighted,kaytan2023gish}, has garnered attention for its smooth, non-step-like behavior and its ability to approximate both linear and ReLU characteristics depending on its parameter. Several empirical studies have demonstrated the superior performance of learnable activation functions in large-scale problems involving tens of thousands of samples or more \cite{apicella2021survey}.

\begin{figure}[htbp!]
    \centering
\includegraphics[width=0.5\linewidth]{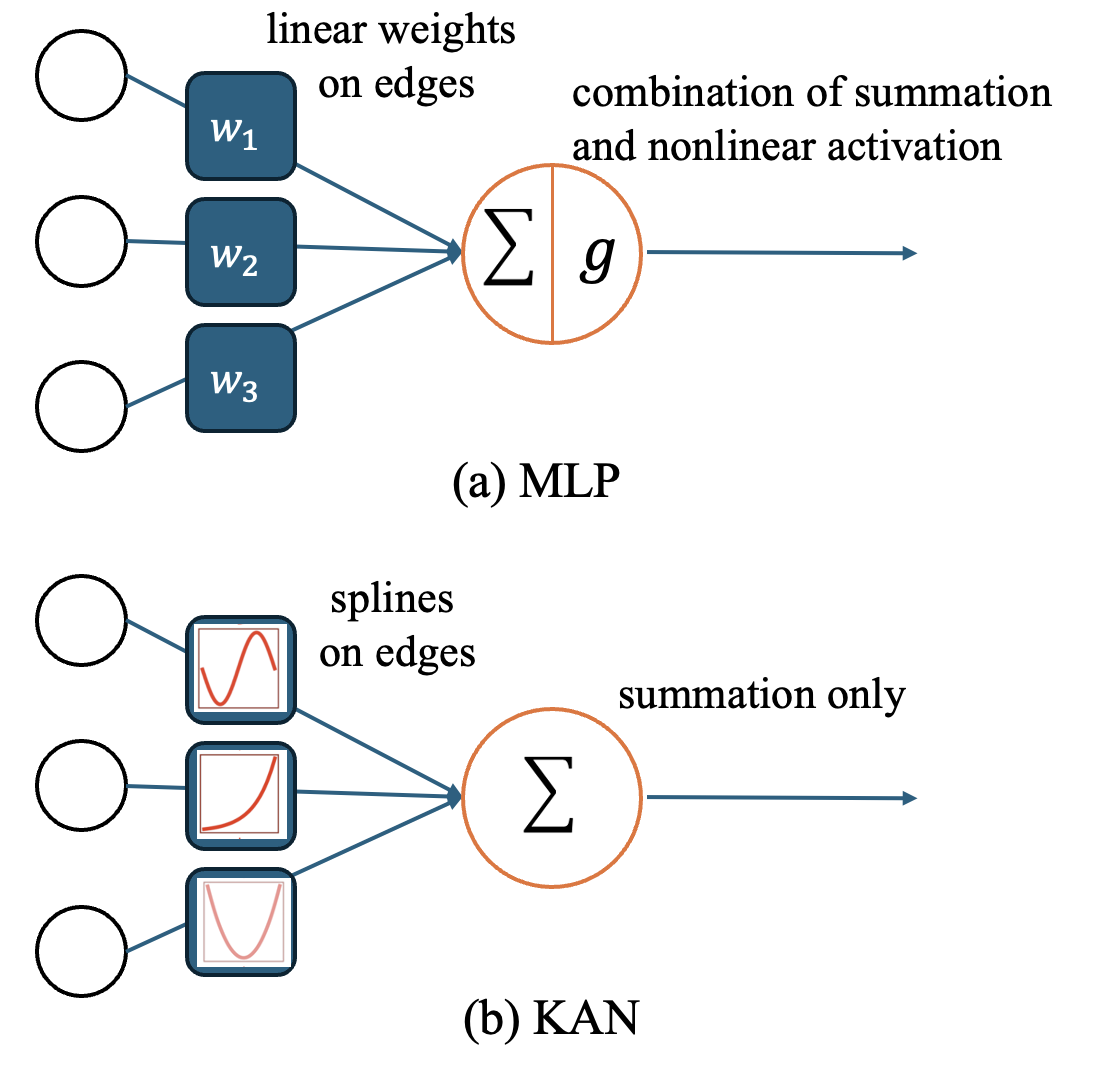}
    \caption{Illustrating the structural differences between Multilayer Perceptrons (MLPs) and Kolmogorov-Arnold Networks (KANs), highlighting that nonlinearities are introduced via the edges or connections between neurons in KANs.}
    \label{fig:atch}
\end{figure}

Recently, Kolmogorov-Arnold Networks (KANs) have been introduced as a novel neural network architecture \cite{liu2024kan}. Unlike traditional MLPs, KANs place learnable activation functions on the connections or edges between neurons, rather than within the neurons themselves. This design is rooted in the mathematical principle that multivariate functions can be decomposed into simpler univariate ones using sums \cite{schmidt2021kolmogorov}.
Figure \ref{fig:atch} illustrates the structural contrast between MLPs and KANs. In MLPs, each neuron performs both summation of weighted inputs and applies a nonlinear activation function, while in KANs, nonlinearity is introduced through the edges of the network themselves.

The recent KAN research \cite{liu2024kan} introduces the concept of a KAN layer, effectively stacking the neurons depicted in Figure \ref{fig:atch}(b), and provides an implementation within a widely-used deep learning framework. This allows for the creation of arbitrarily deep networks through automatic differentiation. In practice, KANs often parameterize their edge activation functions using a combination of a SiLU function and a spline function. Training a KAN then primarily focuses on learning the optimal coefficients for these local B-spline basis functions.

While KANs are designed to utilize more expressive activation functions than standard MLPs with ReLU-like learnable activations, this expressivity comes at the cost of increased parameter count. This raises concerns about their performance in data-sparse scientific and engineering domains, where small sample sizes can hinder effective training.  In many fields, such as medicine and engineering, data collection is often limited to a few tens or hundreds of samples due to high costs and time-consuming procedures \cite{qi2020small,pourkamali2024adaptive}.

Hence, the primary goal of this paper is to provide a comprehensive comparison between MLPs and KANs in low-data scenarios with a few hundred samples. Our key contributions are listed in the following. 
\begin{itemize}
\item Enabling Fair Comparison: Most deep learning libraries implement MLPs with the same activation function applied to all neurons within a layer. Even when using learnable activation functions, neurons in the same hidden layer typically share the same set of parameters. To ensure a fair comparison with KANs, we present a straightforward yet effective technique for designing and implementing MLPs where each neuron in a hidden layer has its own distinct, parameterized activation function. This approach can be applied to networks of any depth or width. Furthermore, we implement a parameterized version of the SiLU activation function to ensure that the activation functions used in MLPs are comparable to those used in KANs.
\item Mathematical Connections: We present a mathematical analysis that elucidates the relationship between MLPs and KANs, showing that KANs can essentially be considered MLPs with activation functions possessing greater flexibility. This analysis underscores the importance of explicitly comparing the number of learnable parameters for both MLPs and KANs in our empirical study.
\item Empirical Evaluation: We conduct experiments on a simulated data set and two real-world classification problems (cancer detection and 3D printer type prediction) to investigate the trade-offs between model complexity (parameter count) and accuracy in data-limited settings. As a key feature of recently introduced KANs is their ability to stack multiple KAN layers, we specifically examine the impact of network depth on testing accuracy across various data splits. Additionally, our experiments reveal that the piecewise polynomial order of splines significantly influences the performance of KANs.
\end{itemize}

The remainder of this paper is organized as follows. In Section \ref{sec:back}, we provide a concise mathematical introduction to MLPs, detail our modification to design MLPs with individual learnable activation functions for each neuron in hidden layers, and explain the underlying transformation in KANs, along with connections to MLPs with trainable activation functions. In Section \ref{sec:syn}, we report experiments on simulated data sets to understand the impact of data size on the predictive accuracy of MLPs and KANs. In Section \ref{sec:real}, we present numerical experiments using real-world data representing complex problems in medicine and engineering. Finally, we conclude this paper with remarks in Section \ref{sec:conc}.

\section{Foundations}\label{sec:back}
A widely adopted neural network architecture for tasks involving structured or tabular data is the Multilayer Perceptron (MLP). These networks are composed of sequentially arranged layers with dense interconnections, providing an effective mechanism for learning multivariate functions. To be formal, imagine a network with $L$ hidden layers between the input and output layers. The $l$-th layer consists of $N_l$ neurons. Also, assume that this network takes an input vector $x\in\mathbb{R}^D$, where $D$ is the number of input features. Moreover, the weight matrix and the bias vector can be written as $W^{(l)}\in\mathbb{R}^{N_l\times N_{l-1}}$ and $b^{(l)}\in\mathbb{R}^{N_l}$, for each layer indexed by $l=1,\ldots,L+1$. The predicted output $f(x)$ is then defined from the input $x$ according to the following equations:
\begin{align}
    &x^{(0)}=x, \nonumber \\
   &x^{(l)}= g^{(l)}\big(W^{(l)}x^{(l-1)}+b^{(l)}\big),\;l=1,\ldots,L, \nonumber \\
    &f(x) = g^{(L+1)}\big(W^{(L+1)}x^{(L)}+b^{(L+1)}\big).\label{eq:nn}
\end{align}
Hence, $f(x)$ takes on a composite or nested form. The number of units in the output layer $N_{L+1}$ and the corresponding activation function $g^{(L+1)}$ depend on the problem at hand. For example, in binary classification problems, a single neuron is typically placed, i.e., $N_{L+1}=1$, and the Sigmoid activation function is used to find the probability that the data point $x$ belongs to one of the two classes \cite{pourkamali2023evaluation}. However, when treating classification problems of more than two classes, the last layer contains one neuron for each class. In this case, we employ the Softmax function to find the categorical distribution for all classes. That is, if we have $C$ classes, then the Softmax function accepts a set of $C$ real-valued numbers $z_1,\ldots,z_C$ and converts them into a valid probability distribution by returning $e^{z_c}/\sum_{c'}e^{z_{c'}}$, for $c=1,\ldots,C$ \cite{ren2020balanced}. 

In contrast, we have more flexibility when choosing activation functions for the $L$ hidden layers, as they produce latent representations. Note that in the equation above, the activation function is applied element-wise. Consequently, the standard implementation of dense layers in popular deep learning libraries like TensorFlow/Keras applies the same activation function to all neurons within a given layer. In other words, every neuron in a layer employs the same nonlinear function to transmit information to the subsequent layer. 

One of the most popular activation functions is the Rectified Linear Unit (ReLU), defined as $\text{ReLU}(z) = \max\{0, z\}$. There are other variants built upon ReLU, such as the Sigmoid Linear Unit (SiLU) \cite{ramachandran2017searching}, also known as the swish function (see \cite{dubey2022activation,jagtap2023important} for a comprehensive list). A key advantage of $\text{SiLU}(z;\beta) = z/(1+e^{-\beta z})$ is its smooth, non-step-like behavior. Its behavior can transition between linear (when $\beta=0$) and ReLU-like (when $\beta$ is sufficiently large). In many cases, such choices, like the value of $\beta$, are predefined hyperparameters and not optimized during training.

\subsection{MLPs with Adaptive Activation Functions}\label{sec:mlp}
One approach to enhancing the performance and adaptability of MLPs is to treat the parameters within activation functions as trainable during the learning process. This involves incorporating these parameters into the computation graph, allowing the calculation of gradients of the loss function with respect to them. As a result, these activation function parameters, such as the optimal value of $\beta$ for SiLU, can be learned alongside the network weights. 

While several empirical studies \cite{apicella2021survey,biswas2022erfact,kiliccarslan2024parametric} have demonstrated the advantages of adaptive activation functions over fixed ones in MLPs, particularly in large-scale data settings like computer vision, it is common to assume that all neurons within a layer share the same activation function parameters. In other words, in Equation \eqref{eq:nn}, $g^{(l)}$ is identical for all $N_l$ neurons in the $l$-th layer. However, for a fair comparison with KANs, where each connection has its own unique activation function, it is crucial to allow each activation function in MLPs to operate independently, increasing their expressivity. 

To achieve this goal, we utilize the built-in concatenation layer available in  deep learning libraries, enabling us to seamlessly merge the outputs of neurons with distinct activation functions. Let $W_i^{(l)}$ denote the $i$-th row of the weight matrix $W^{(l)}$, and let $b_i^{(l)}$ represent the $i$-th element of $b^{(l)}$. Then, each neuron in the $l$-th layer computes its output as follows:
\begin{equation}
a_i^{(l)}=g_i^{(l)}\big(W_i^{(l)}x^{(l-1)}+b_i^{(l)}\big),\;i=1,\ldots,N_l,
\end{equation}
where $g_i^{(l)}$ is the activation function specific to the $i$-th neuron, with its own learnable parameter $\beta_i$. The outputs from all $N_l$ neurons are then concatenated to form the final output of the $l$-th hidden layer:
\begin{equation}
x^{(l)}=\text{concatenate}\big([a_1^{(l)}, \ldots, a_{N_l}^{(l)}]\big).
\end{equation}
This approach offers significant flexibility, allowing us to design and implement MLPs of any depth and width. Each neuron has its own unique activation function, and the entire process leverages standard automatic differentiation within existing deep learning libraries.

\subsection{Kolmogorov-Arnold Networks (KANs)}
KANs depart from the traditional MLP architecture by placing learnable functions directly on the network's edges (connections between neurons), rather than within the neurons themselves. Consequently, the primary role of each neuron is to simply sum its incoming signals, without applying any additional nonlinearities. This design aims to integrate the nonlinear transformation directly into the weighting mechanism of the edges, rather than separating linear weighting and nonlinear activation as in conventional MLPs.

To formalize this, let us describe the transformation performed by the $l$-th layer of a KAN, drawing parallels to Equation \eqref{eq:nn}. Recall that the incoming signal $x^{(l-1)}$ has $N_{l-1}$ dimensions and the output of the $l$-th layer has $N_l$ elements. Let  $g_{j,i}^{(l)}(\cdot)$ denote a univariate function that operates on the $i$-th dimension of $x^{(l-1)}$
to be used for calculating the $j$-th element of the output. Then, we have the following transformation performed by the $l$-th KAN layer:
\begin{equation}
x^{(l)}=\begin{bmatrix}g_{1,1}^{(l)}(\cdot) & \ldots & g_{1,N_{l-1}}^{(l)}(\cdot)\\ g_{2,1}^{(l)}(\cdot) & \ldots & g_{2,N_{l-1}}^{(l)}(\cdot)\\
    \vdots &  & \vdots\\g_{N_l,1}^{(l)}(\cdot) & \ldots & g_{N_l,N_{l-1}}^{(l)}(\cdot) \end{bmatrix}x^{(l-1)}.\label{eq:mat}
\end{equation}
Consequently, the transformation performed by each layer remains expressible as a matrix-vector multiplication. However, a crucial aspect of designing KANs lies in the selection of activation functions. For simplicity, let us omit all superscripts and subscripts in the following discussion. It has been proposed that each activation function in Equation \eqref{eq:mat} can be represented as the weighted sum of SiLU and a spline function:
\begin{equation}
g(x) = w_b\text{SiLU}(x)+w_s \text{spline}(x), \label{eq:act}
\end{equation}
where $\text{SiLU}(x)=x/(1+e^{-x})$ represents the Sigmoid Linear Unit function with $\beta=1$ and $\text{spline}(x)=\sum_{i}c_iB_i(x)$ is a linear combination of B-splines \cite{unser2019representer,bohra2020learning}. Thus, the training process involves learning the optimal values of $c_i$, $w_b$, and $w_s$.
This approach enables high expressivity by utilizing rich activation functions that go beyond the popular ReLU function typically used in MLPs. Moreover, both the degree of each spline (spline order) and the number of splines used for each function are hyperparameters of the KAN architecture.

\subsection{Bridging MLPs and KANs}

While KANs clearly utilize activation functions with greater flexibility than standard MLPs employing functions like SiLU, it is worth questioning whether the KAN architecture is truly novel. Comparing Equations \eqref{eq:nn} and \eqref{eq:mat} reveals that layer transformations in both architectures can be expressed as matrix-vector multiplications. However, a key distinction lies in the order of operations. MLPs first compute a weighted sum of inputs from the previous layer, followed by a nonlinear activation function. In contrast, KANs apply nonlinear transformations to the inputs first, then perform a weighted sum.

To explain this, we rewrite Equation \eqref{eq:act}:
\begin{equation}
    g(x)=\begin{bmatrix} w_b & w_sc_1& w_sc_2 &\ldots
    \end{bmatrix}\begin{bmatrix}\text{SiLU}(x)\\ B_1(x)\\B_2(x)\\\vdots\end{bmatrix}.
\end{equation}
Therefore, the activation function used on KAN edges can be viewed as a nonlinear transformation followed by a weighted sum. This is similar to MLPs, except for the order of these operations. From a practical standpoint, the critical question is whether this difference in ordering, along with the increased number of trainable parameters in KANs, leads to higher accuracy levels compared to MLPs with trainable activation functions, particularly in low-data regimes. The remainder of this paper will focus on investigating this question.
\section{Performance Evaluation: Synthetic Data}\label{sec:syn}
\begin{figure}[htbp!]
    \centering
\includegraphics[width=0.7\linewidth]{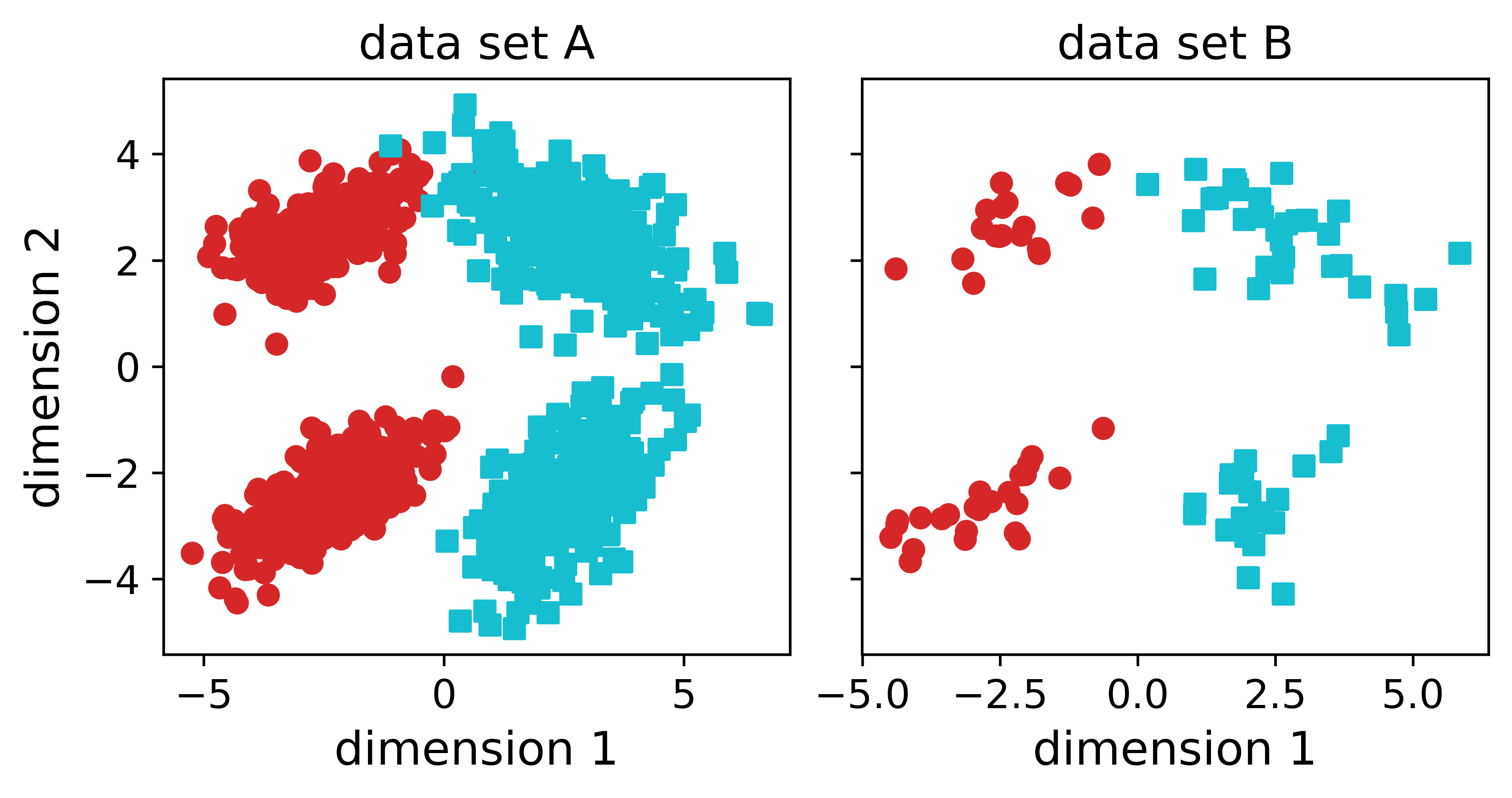}
    \caption{Visualization of the two simulated data sets with varying sample sizes (1,000 vs.~100 data points).}
\label{fig:dataviz}
\end{figure}

In this section, we commence our comparative study between MLPs and KANs using a two-dimensional simulated data set. This data set encompasses two classes, each consisting of two clusters. Our objective is to evaluate their performance on a separable yet relatively complex data set. Figure \ref{fig:dataviz} illustrates the two data sets central to our analysis: data set A (1,000 samples) and data set B (100 samples). This enables us to directly understand the impact of data size on the performance of KANs, which possess higher degrees of freedom due to the choice of activation functions; see Equation \eqref{eq:act}. For consistency, the default values of grid size 3 and spline order 3 are employed for KANs in this paper, with the exception of Section \ref{sec:hyper} where we investigate the impact of hyperparameter adjustments.

Furthermore, to ensure a fair comparison, we utilize the parameterized SiLU activation for the MLP architecture with the learnable parameter $\beta$. As previously discussed in Section \ref{sec:mlp}, each neuron in this MLP architecture possesses its own set of trainable parameters, achieved through the concatenation process. For model fitting, we consistently use 20 epochs and a fixed learning rate of 0.05 across all experiments in this paper.

Our comparative analysis focuses on varying network depth $L$ while maintaining a fixed width (number of neurons per hidden layer) to ensure a compact network for evaluating performance in low-data regimes and align with the KAN paper's focus on the KAN layer's stacking potential for deeper models. All experiments use a width of 2 and consider depths of 1, 2, and 3, employing a 70-30 train-test split and measuring accuracy on the testing set, repeated 25 times to account for the impact of data splits and weight initializations. 

For our visualizations, we select violin plots, which provide a richer understanding of the data distribution compared to traditional box plots. The width of each violin at a particular value reflects the concentration of data points around that value, while the vertical axis represents the probability density. Additionally, a tick mark within each violin pinpoints the median of the associated evaluation metric. 

\begin{figure}[htbp!]
    \centering
\includegraphics[width=0.7\linewidth]{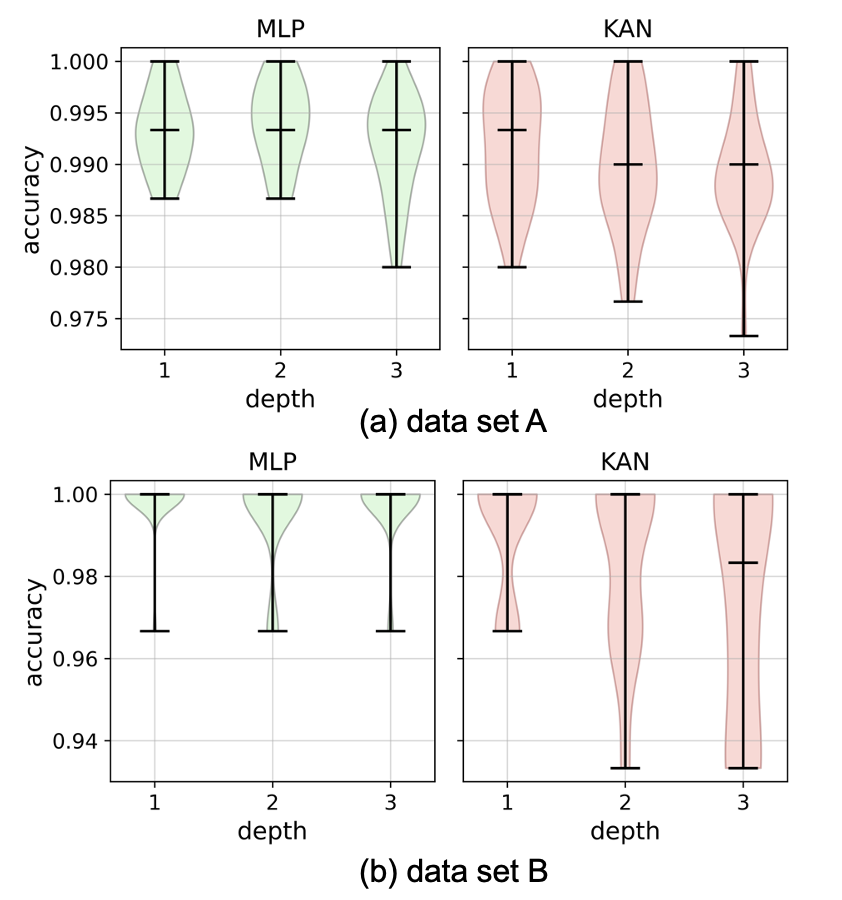}
\caption{Classification accuracy comparison: MLPs vs. KANs on simulated data sets A and B.}
    \label{fig:syn_acc}
\end{figure}

As shown in Figure \ref{fig:syn_acc}(a), both models perform well on data set A due to its separability, resulting in classification accuracies near 1. However, it is noteworthy that both models experience a slight decrease in accuracy as depth increases. Interestingly, this decrease is slightly more pronounced in KANs. For instance, KANs with a depth of 3 reach a minimum accuracy of 0.973, the lowest observed accuracy level. Furthermore, the median accuracy for KANs with depth 3 is 0.990, compared to approximately 0.993 for all MLP depths. 

Furthermore, Figure \ref{fig:syn_acc}(b) reveals a larger performance gap between MLPs and KANs on the smaller data set (data set B). While MLPs maintain accuracy levels comparable to the previous data set (e.g., median values near 1), the performance of KANs noticeably degrades in this small-scale scenario. For instance, KANs with depth 3 have a median accuracy of 0.983 and a minimum accuracy of 0.933, a significant drop compared to the minimum of 0.973 observed for data set A. When comparing the distribution of results across all depths, MLPs show a clear advantage over KANs. 

\begin{figure}[htbp!]
    \centering
\includegraphics[width=0.6\linewidth]{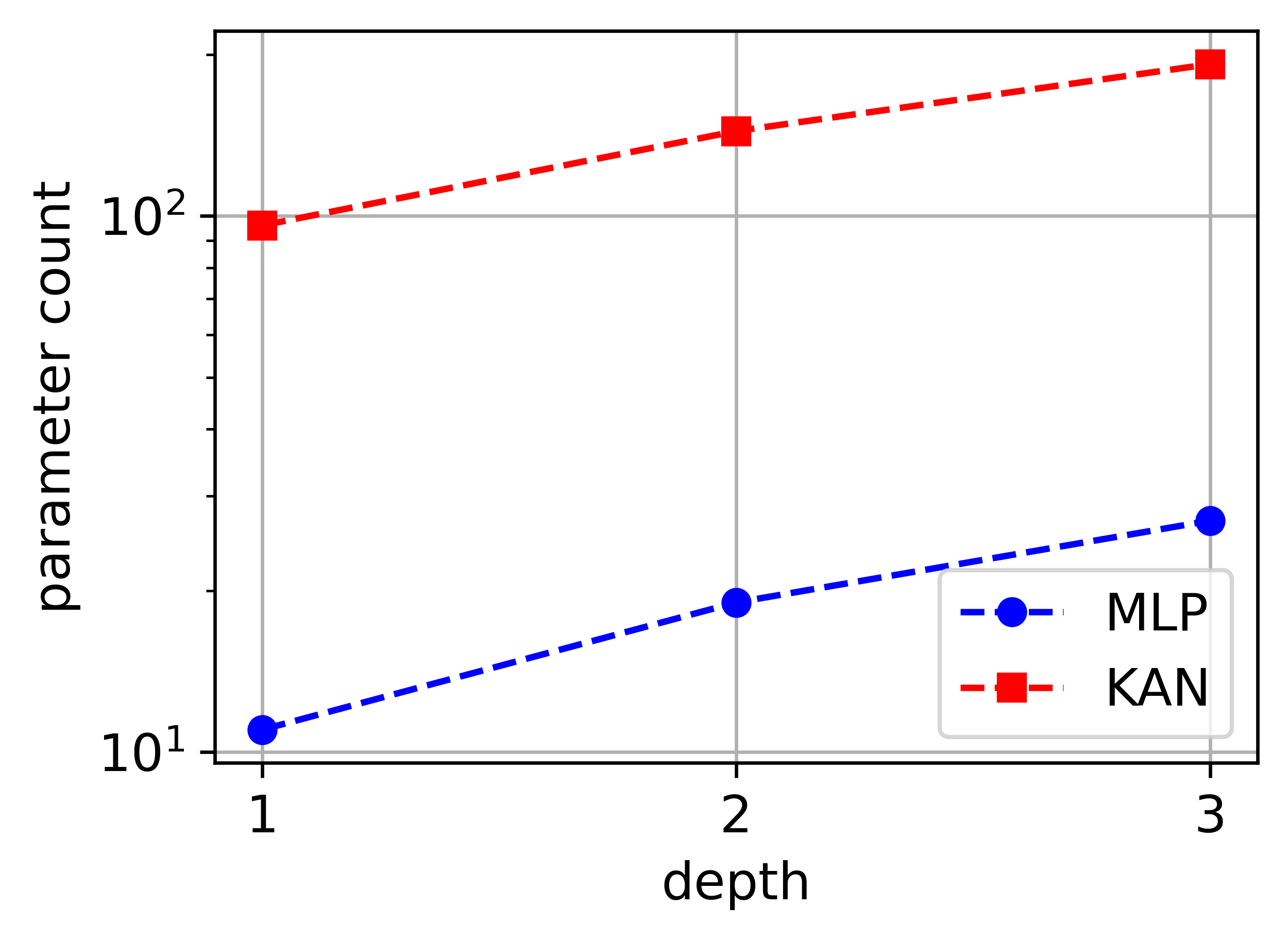}
    \caption{Parameter count comparison: a substantial disparity in the total number of learnable parameters is observed between KANs and MLPs on simulated data.}
    \label{fig:syn_count}
\end{figure}

To gain deeper insights into the behavior of MLPs and KANs in relation to depth, we present the total number of learnable parameters in Figure \ref{fig:syn_count}. This reveals striking differences between the two models due to their degrees of freedom. KANs, using more complex spline functions compared to the parameterized SiLU in MLPs, have an order of magnitude more parameters to learn. For example, at depth 3, MLPs have 27 learnable parameters while KANs have 192. This suggests that even with distinct activation functions per neuron, the parameterization used by MLPs provides sufficient flexibility to effectively classify this data set with its two clusters per class. Hence, this analysis highlights that MLPs have a distinct advantage over KANs when dealing with limited labeled data.

\section{Real-World Case Studies}\label{sec:real}

\subsection{Cancer Detection}
In this case study, we use the breast cancer wisconsin data set from scikit-learn, which is a well-known data set in machine learning and medical research. It consists of diagnostic information about breast cancer, including input features derived from digitized images of fine needle aspirates (FNAs) of breast masses. The data set contains 569 samples and each sample is described by 30 real-valued input features, thus $D=30$. There are 212 malignant and 357 benign samples in the data set, providing a somewhat balanced view of both malignant and benign cases, with the goal of training machine learning models to classify the diagnosis based on the input features.

As detailed in the previous section, we maintain a constant network width of 2 while varying the depth $L$ (1, 2, and 3). We also ensure consistency by using the same number of epochs and learning rate values. To assess classification accuracies, we conduct 25 independent experiments, each with its own train-test split and weight initialization. Our goal is to compare the performance of MLPs and KANs across these 25 repetitions using violin plots.

\begin{figure}[htbp!]
    \centering
\includegraphics[width=0.7\linewidth]{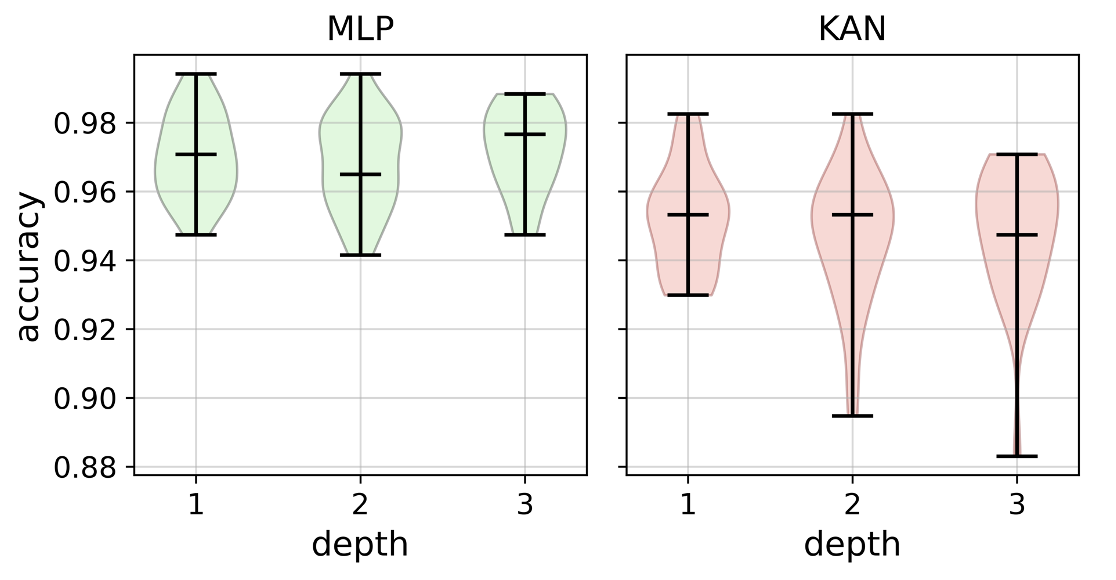}
    \caption{Evaluating the effect of depth (number of hidden layers) on classification accuracy: a comparative study of MLPs and KANs using the cancer detection data set with 569 samples.}
    \label{fig:cancer_acc}
\end{figure}

Figure \ref{fig:cancer_acc} demonstrates that MLPs significantly outperform KANs in this cancer detection problem. The maximum and minimum accuracy levels for MLPs across all three depths are approximately 0.99 and 0.94, respectively. Therefore, our implementation of MLPs with individually trainable parameters in their SiLU activation functions aligns with existing neural network implementations on this popular data set, which have achieved high accuracies approaching 0.99 \cite{alshayeji2022computer}.

In contrast, the maximum and minimum accuracy values for KANs are around 0.98 and 0.88, respectively. This wider range of accuracy values, approximately twice that of MLPs, suggests lower overall accuracy and potentially less reliability in their predictions. Similar results are observed when comparing median accuracy levels. For instance, at depth 3, the median accuracy of MLPs is 0.976, while KANs have a substantially lower median value of 0.947. Importantly, MLPs' performance remains fairly consistent across the three depths, suggesting that hyperparameter tuning may be more straightforward compared to KANs.

\begin{figure}[htbp!]
    \centering
\includegraphics[width=0.6\linewidth]{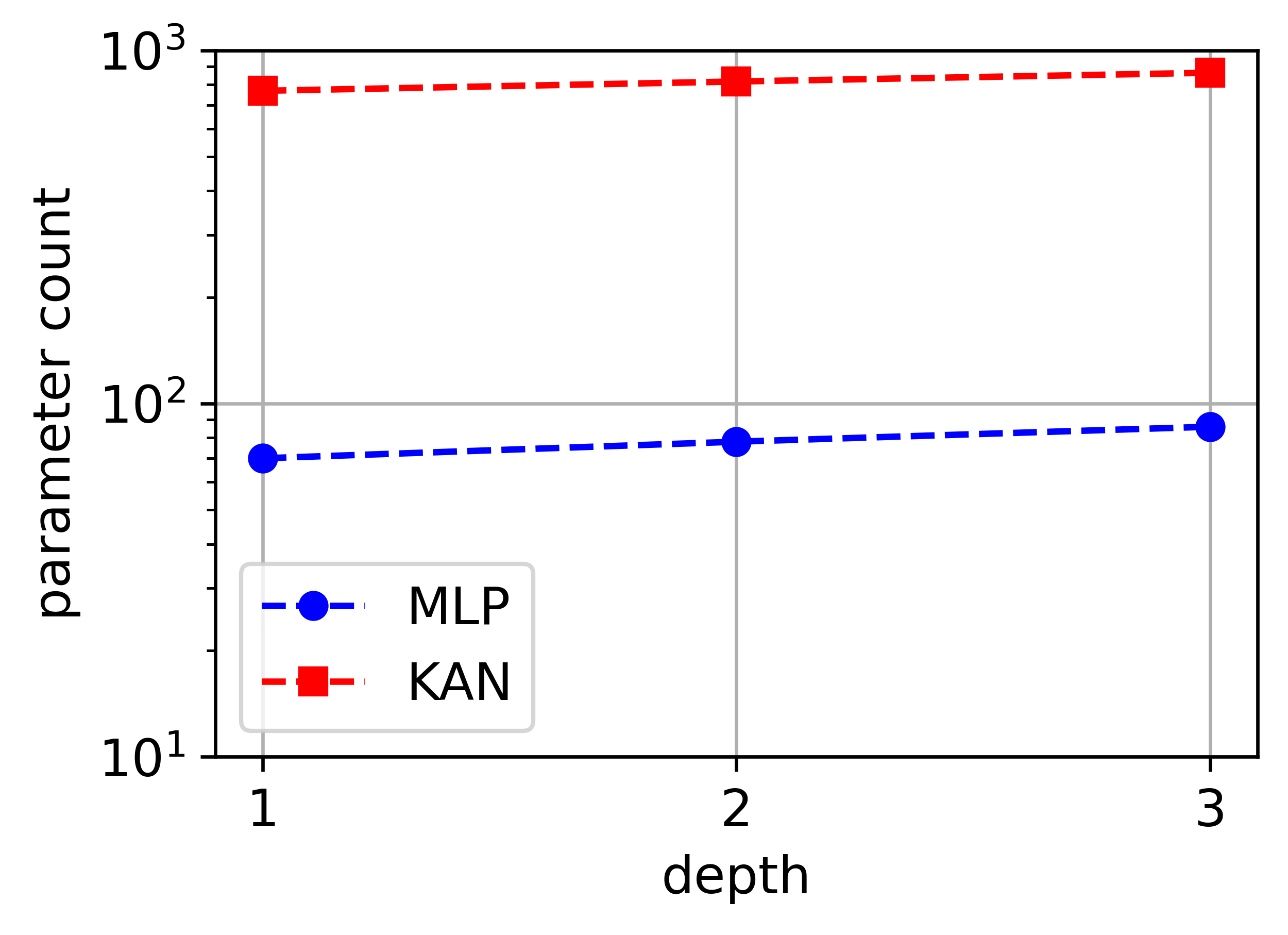}
    \caption{On the cancer data set, KANs have a substantially higher number of learnable parameters compared to MLPs.}
    \label{fig:cancer_count}
\end{figure}

To further analyze the impact of activation function choice on model complexity, we present the total number of learnable parameters for MLPs and KANs in Figure \ref{fig:cancer_count}, particularly considering the 30-dimensional input. The figure reveals a stark contrast: KANs generally possess an order of magnitude more parameters than MLPs. Despite this increased complexity, KANs underperform in this real-world cancer detection task. This suggests that MLPs, even with individual parameterized activation functions, achieve sufficient complexity for accurate classification using fewer parameters. In this case, the simpler MLP architecture appears to be more effective at learning from the 30 input features, highlighting the potential advantages of parameter efficiency in low-data regimes. 

\subsection{3D Printer Type Prediction}
In fused filament fabrication (FFF), the mechanical properties of printed parts are influenced not only by printing parameters but also by the specific 3D printer used. Variations in hardware and firmware across printer models lead to differences in material deposition, movement precision, and temperature control, impacting factors like interlayer adhesion and ultimately the final part's strength and surface quality \cite{nasrin2024application}. 

In this section, we evaluate classification models based on MLPs and KANs with varying depths for identifying the 3D printer used to manufacture a given part. Using a data set \cite{braconnier2020processing} comprising tensile properties of parts printed on three different printers (MakerBot Replicator 2X, Ultimaker 3, and Zortrax M200) with varying printing parameters, our models aim to predict the printer type based on 7 input features: tensile strength, elastic modulus, elongation at break, extrusion temperature, layer height, print bed temperature, and print speed. This approach seeks to capture the subtle relationships between printing process, part properties, and the specific printer used, which can be viewed as a three-class classification problem with $D=7$ input features and 104 samples. Thus, this section aims to evaluate the performance of MLPs and KANs in scenarios with severely limited data, a common challenge in many experimental settings.

\begin{figure}[htbp!]
    \centering
\includegraphics[width=0.7\linewidth]{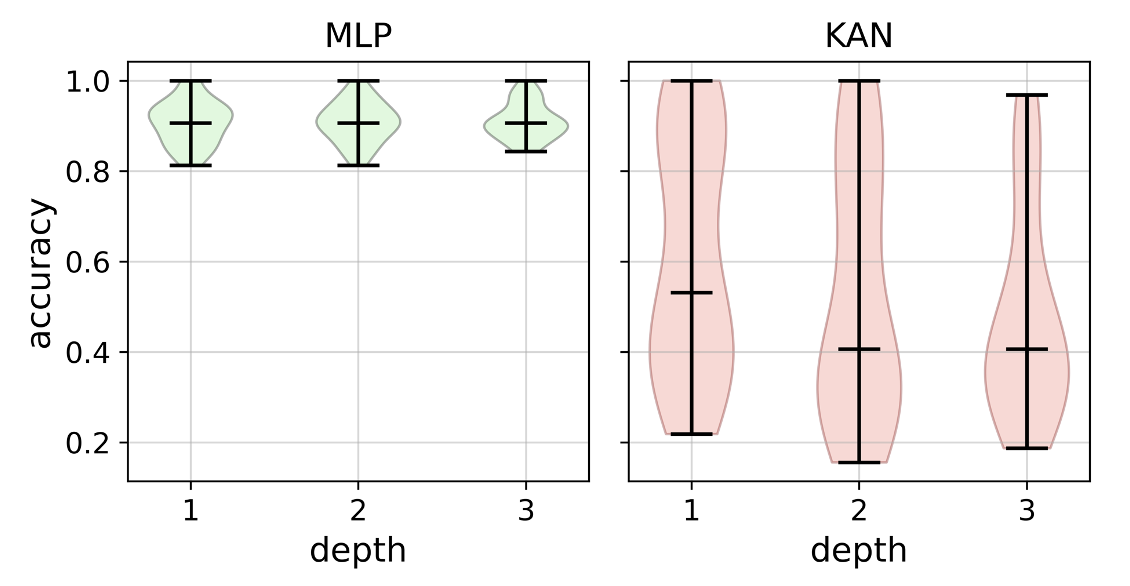}
    \caption{Evaluating the effect of depth (number of hidden layers) on classification accuracy: a comparative study of MLPs and KANs using the 3D printer type prediction data set with 104 samples.}
    \label{fig:printer_acc}
\end{figure}

Figure \ref{fig:printer_acc} presents classification accuracy results on this data set across 25 repetitions, each utilizing a 70-30 train-test split ratio, to capture the effects of weight initialization and other stochastic factors in training neural network models. Overall, MLPs demonstrate strong performance across all three depth values. For instance, with a depth of 3, the maximum, median, and minimum accuracies are 1, 0.906, and 0.844, respectively. These results highlight the effectiveness of MLPs, especially considering this is a three-class classification problem where a random classifier would yield accuracies closer to 0.333, significantly lower than our observed results.

On the other hand, we observe a more pronounced accuracy drop in KANs compared to the previous cancer detection data set. While KANs can achieve accuracies at or near 1 in some repetitions, there are instances where accuracy falls below 0.333, the baseline for a random classifier. Fortunately, the median values for KANs at depths 1, 2, and 3 are 0.531, 0.406, and 0.406, respectively, all surpassing the baseline. Nevertheless, MLPs consistently outperform KANs in this scenario.

\begin{figure}[htbp!]
    \centering
\includegraphics[width=0.6\linewidth]{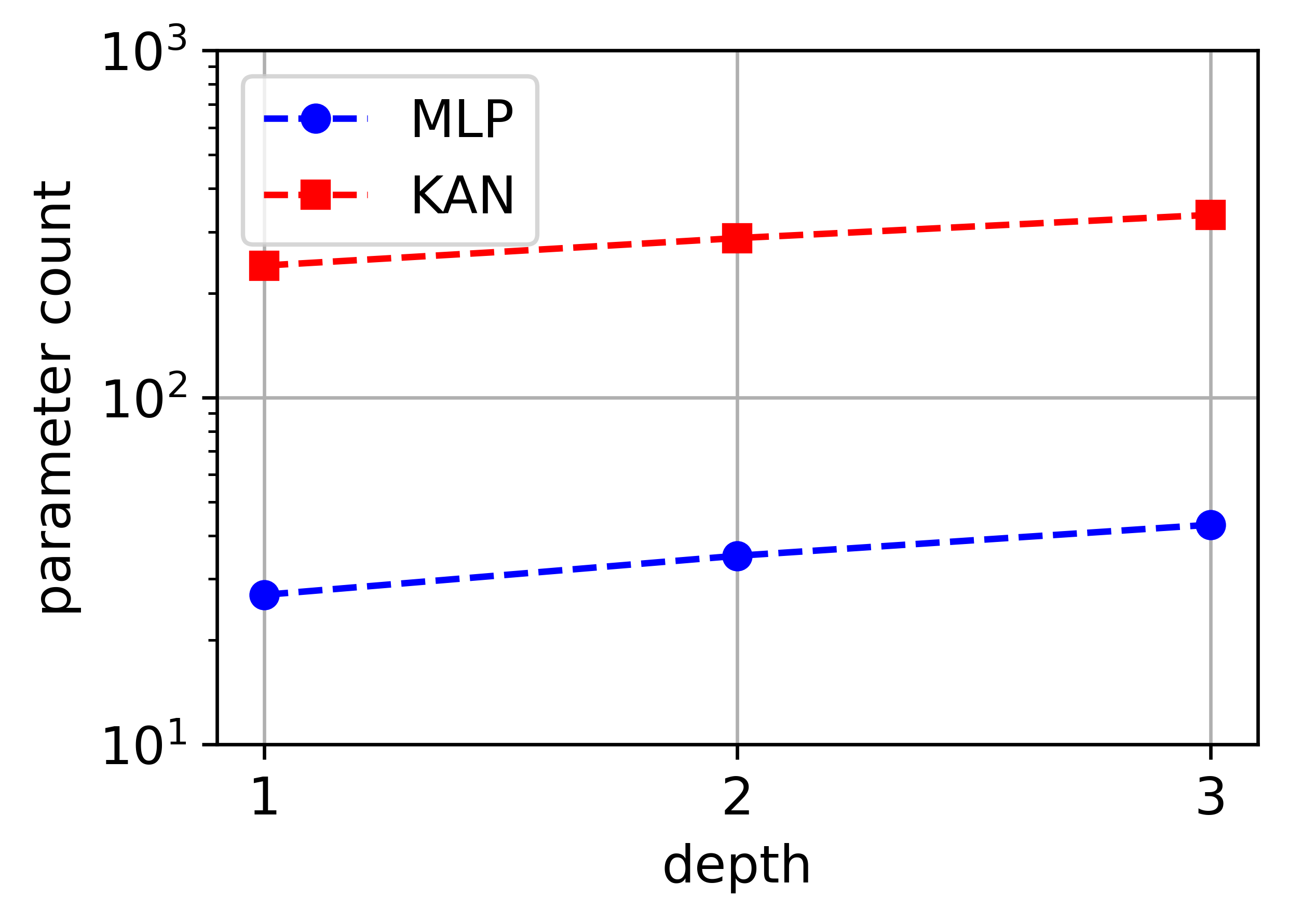}
    \caption{On the 3D printer type prediction data set, MLPs possess significantly fewer learnable parameters, a number that remains lower than the total number of samples available in this data set.}
    \label{fig:printer_count}
\end{figure}

Similar to previous cases, we also report the total number of learnable parameters on the 3D printer type prediction data set in Figure \ref{fig:printer_count}. Again, we observe that MLPs have approximately an order of magnitude fewer parameters to learn, which is significant given the small sample size. Specifically, the total number of learnable parameters for MLPs at depths 1, 2, and 3 are 27, 35, and 43, respectively. These values are below the sample size, which may explain the consistent good performance of MLPs on the test set across different repetitions.

\subsection{Dependence of KANs on the Polynomial Order of Activations}\label{sec:hyper}

In this section, we delve deeper into the performance of KANs on these two real-world data sets to understand how the complexity of the activation functions influences their behavior. Recall that each activation function in KANs is parameterized as a B-spline, and a crucial hyperparameter is the polynomial order of these splines. An order of 1 results in an activation function similar to ReLU, i.e., a piecewise linear function, while higher orders provide increasing degrees of nonlinearity.

The default spline order in the KAN implementation is 3, offering a reasonable balance of nonlinearity. To gain further insight into the impact of this choice in low-data regimes, we consider networks of depth 2 and width 2, but we vary the spline order from 1 to 5. 

\begin{figure}[htbp!]
    \centering
\includegraphics[width=0.9\linewidth]{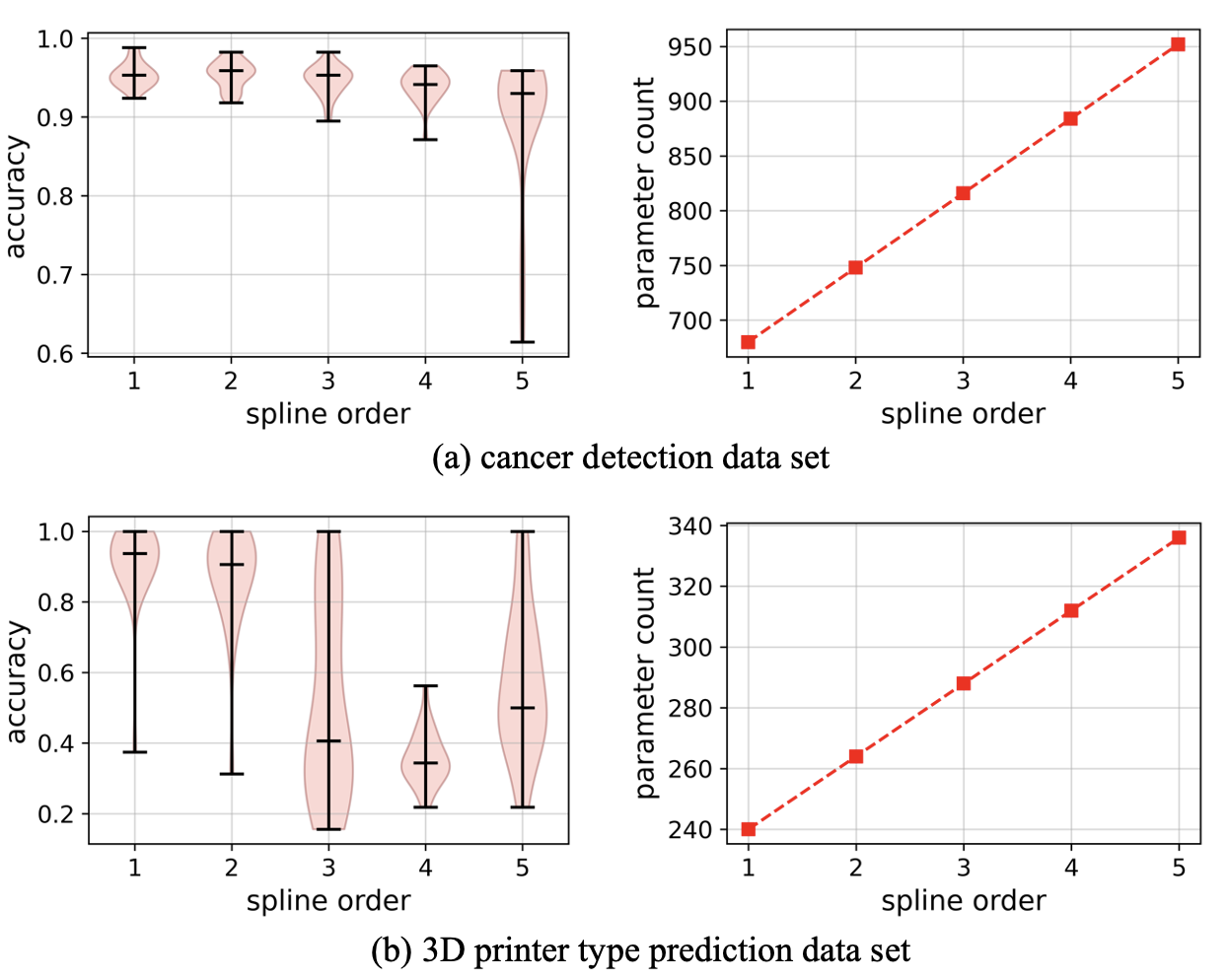}
    \caption{Investigating the impact of spline order on both the accuracy of KANs and the number of learnable parameters, utilizing two real-world data sets: (a) cancer detection and (b) 3D printer type prediction.}
    \label{fig:order}
\end{figure}

Figure \ref{fig:order}(a) reveals an interesting trend: in most cases, the accuracy of KANs on the cancer detection data set decreases as the spline order increases. The highest median accuracy is achieved at order 2 (0.959), while the lowest is at order 5 (0.929). However, MLPs with parameterized swish activations per neuron still outperform the best KAN configuration in this experiment. The median accuracy for MLPs with depth 2 reaches 0.965, while maintaining an order of magnitude fewer learnable parameters. This demonstrates that MLPs with individualized parameterized activations can achieve higher accuracy with significantly fewer parameters, a crucial advantage in low-data scenarios.

Furthermore, we investigate the impact of spline order on the 3D printer type prediction data set, which has a substantially smaller sample size and involves a trinary (three-class) classification problem instead of binary. In this case, Figure \ref{fig:order}(b) demonstrates that even a spline order of 1 can lead to a noticeable number of instances where accuracy falls below 0.8, the minimum value observed for MLPs in the previous section. While all spline orders except for 4 can achieve high accuracies close to 1, a serious concern is the potential for substantial performance degradation due to data splits and other stochastic factors in each repetition. Moreover, even with a spline order of 1, KANs have a much higher number of learnable parameters compared to MLPs. This suggests that more compact MLPs can achieve superior accuracy levels in this scenario.

\section{Conclusions and Future Directions}\label{sec:conc}

While Kolmogorov-Arnold Networks (KANs) offer an intriguing alternative to Multilayer Perceptrons (MLPs) by replacing linear weights with highly expressive activation functions, our findings highlight their notable performance degradation in low-data regimes compared to MLPs. As our algorithmic comparison revealed, the primary strength of KANs lies in their choice of activation functions because they implicitly incorporate a linear weighting mechanism similar to MLPs. This underscores the critical role of activation function complexity in neural network performance. While complex activation functions with greater flexibility might seem like the obvious choice, our findings suggest that simpler alternatives such as the SiLU can provide adequate capacity for some practical applications, particularly in scenarios with limited data availability.

Additionally, our research has revealed that individually parameterized neurons within hidden layers can derive advantages from utilizing independent, individualized activation functions. This approach does not compromise accuracy and opens new avenues for enhancing the predictive power of smaller networks on small-scale data sets. Implementing such MLPs with fully adaptive activation functions is straightforward in popular deep learning libraries like TensorFlow/Keras and PyTorch using concatenation layers. It is imperative to include these MLPs in future comparative studies, as existing benchmarks focusing on fixed-shape activation functions, e.g., \cite{poeta2024benchmarking}, may not provide a fair comparison. This paves the way for exploring novel activation functions that offer controlled nonlinear transformations for analyzing complex data.

Several recent works have explored activation functions beyond splines in KANs, which can also be applied to MLPs with individualized activations. For example, wavelet functions can capture both high-frequency and low-frequency components of input data \cite{bozorgasl2024wav}. Another promising direction is developing adaptive algorithms for selecting activation functions per neuron in MLPs, considering a predefined space including splines, wavelets, Chebyshev polynomials, and others \cite{yang2024activation,shukla2024comprehensive}. Such an approach could factor in sample size and data complexity measures, proving especially beneficial in low-data scenarios.

Finally, future comparative studies should also investigate the sensitivity of MLPs and KANs to their hyperparameters. Our experiments demonstrated that MLPs are relatively insensitive to network depth, whereas KAN performance can significantly degrade with increasing depth or spline order. A thorough comparative analysis necessitates a comprehensive examination of other hyperparameters like learning rate, number of epochs, and grid size, to name a few.

\bibliographystyle{ieeetr}
\bibliography{sn-bibliography}

\end{document}